\pdfoutput=1
\documentclass[11pt]{article}
\usepackage[table,dvipsnames]{xcolor}
\usepackage[]{acl}

\usepackage{times}
\usepackage{latexsym}

\usepackage[T1]{fontenc}
\usepackage[utf8]{inputenc}

\usepackage{microtype}

\usepackage{inconsolata}

\usepackage{graphicx}
\usepackage{inconsolata}
\usepackage{array,multirow}

\newlength{\myMheight}
\usepackage{booktabs}
\usepackage{tabularx}
\usepackage{soul}
\usepackage[most]{tcolorbox}
\usepackage{listings}
\usepackage{collcell}
\usepackage{extdash} % needed for suppressing linebreaks at hyphens in a table
\usepackage{fontawesome5}

\definecolor{Gray}{gray}{0.9}
\definecolor{cb-blue-green} {RGB}{ 0,  073,  073}
\definecolor{cb-green-sea}  {RGB}{ 0, 146, 146}
\definecolor{cb-rose}       {RGB}{255, 109, 182}
\definecolor{cb-salmon-pink}{RGB}{255, 182, 119}
\definecolor{cb-purple}     {RGB}{ 73,   0, 146}
\definecolor{cb-blue}       {RGB}{ 0, 109, 219}
\definecolor{cb-lilac}      {RGB}{182, 109, 255}
\definecolor{cb-blue-sky}   {RGB}{109, 182, 255}
\definecolor{cb-blue-light} {RGB}{182, 219, 255}
\definecolor{cb-burgundy}   {RGB}{146,   0,   0}
\definecolor{cb-brown}      {RGB}{146,  73,   0}
\definecolor{cb-clay}       {RGB}{219, 209,   0}
\definecolor{cb-green-lime} {RGB}{ 36, 255,  36}
\definecolor{cb-yellow}     {RGB}{255, 255, 109}
\definecolor{cb-grey}       {RGB}{233, 233, 233}

\definecolor{lt-red}       {RGB}{186, 33, 33}
\definecolor{lt-blue}       {RGB}{0, 0, 255}
\definecolor{lt-green}       {RGB}{0, 128, 0}
\definecolor{lt-pale}       {RGB}{61, 122, 122}

\lstdefinestyle{mystyle}{   
    commentstyle=\color{lt-pale},
    keywordstyle=\color{lt-green},
    numberstyle=\tiny\color{cb-grey},
    stringstyle=\color{lt-red},
    basicstyle=\fontsize{8}{10.8}\selectfont\ttfamily,
    breakatwhitespace=false,         
    breaklines=true,                 
    captionpos=b,               
    keepspaces=true,                 
    showspaces=false,                
    showstringspaces=false,
    showtabs=false,
}

\usepackage{tikz}
\usepackage{collcell}
\usepackage{pgf}
\newcommand*{\MinNumber}{0.0}%
\newcommand*{\MidNumber}{60.0} %
\newcommand*{\MaxNumber}{100.0}%

\newcommand{\ApplyGradient}[1]{%
        \ifdim #1 pt > \MidNumber pt
            \pgfmathsetmacro{\PercentColor}{max(min(100.0*(#1 - \MidNumber)/(\MaxNumber-\MidNumber),100.0),0.00)} %
            \hspace{-0.33em}\colorbox{SeaGreen!\PercentColor!Goldenrod!50}{#1}
        \else
            \pgfmathsetmacro{\PercentColor}{max(min(100.0*(\MidNumber - #1)/(\MidNumber-\MinNumber),100.0),0.00)} %
            \hspace{-0.33em}\colorbox{Red!\PercentColor!Goldenrod!50}{#1}
        \fi
}

\newcommand{\CG}[3]{% #1 is MaxValue, #2 is MinValue, #3 is the cell value
  \def\MaxValueCmd{#1}
  \def\MinValueCmd{#2}
  \pgfmathsetmacro{\MidValueCmd}{(\MaxValueCmd + \MinValueCmd) / 2}
  
  \ifdim #3 pt > \MidValueCmd pt
    \pgfmathsetmacro{\PercentColorCmd}{max(min(100.0*(#3 - \MidValueCmd)/(\MaxValueCmd-\MidValueCmd),100.0),0.00)}
    \hspace{-0.33em}\colorbox{SeaGreen!\PercentColorCmd!Goldenrod!50}{#3}
  \else
    \pgfmathsetmacro{\PercentColorCmd}{max(min(100.0*(\MidValueCmd - #3)/(\MidValueCmd-\MinValueCmd),100.0),0.00)}
    \hspace{-0.33em}\colorbox{Red!\PercentColorCmd!Goldenrod!50}{#3}
  \fi
}

\newcommand{\CGold}[2]{% #1 is MaxValue, #2 is the cell value
  \def\MaxValueCmd{#1}
  \pgfmathsetmacro{\MidValueCmd}{\MaxValueCmd/1.5}
  \pgfmathsetmacro{\MinValueCmd}{\MaxValueCmd/2}
  \ifdim #2 pt > \MidValueCmd pt
    \pgfmathsetmacro{\PercentColorCmd}{max(min(100.0*(#2 - \MidValueCmd)/(\MaxValueCmd-\MidValueCmd),100.0),0.00)}
    \hspace{-0.33em}\colorbox{SeaGreen!\PercentColorCmd!Goldenrod!50}{#2}
  \else
    \pgfmathsetmacro{\PercentColorCmd}{max(min(100.0*(\MidValueCmd - #2)/(\MidValueCmd-\MinNumberCmd),100.0),0.00)}
    \hspace{-0.33em}\colorbox{Red!\PercentColorCmd!Goldenrod!50}{#2}
  \fi
}

\newcolumntype{R}{>{\collectcell\ApplyGradient}c<{\endcollectcell}}

\tcbuselibrary{skins}
\newcommand{\improvedParbox}[1]{%
  \begin{tcolorbox}[
    enhanced,
    colback=white,
    colframe=gray!50!black,
    boxrule=0.5pt,
    arc=1mm,
    left=3mm,
    right=3mm,
    top=2mm,
    bottom=2mm,
    boxsep=0pt,
    width=\columnwidth
  ]
    #1
  \end{tcolorbox}
}

\usepackage{adjustbox}

\def\cquad{\hskip0.8em\relax}

\title{{\huge $\tau$}~\textsc{tau-eval}: A Unified Evaluation Framework for Useful and Private \\Text Anonymization}

\author{Gabriel Loiseau$^{1,2}$ \cquad Damien Sileo$^{2}$ \cquad Damien Riquet$^{1}$ \cquad Maxime Meyer$^{1}$ \cquad Marc Tommasi$^{2}$ \\ $^{1}$Hornetsecurity, Hem, France \\ $^2$Univ. Lille, Inria, CNRS, Centrale Lille, UMR 9189 - CRIStAL, F-59000 Lille, France \\
\texttt{gabriel.loiseau@inria.fr}}

\begin{document}
\maketitle
\begin{abstract}
Text anonymization is the process of removing or obfuscating information from textual data to protect the privacy of individuals. This process inherently involves a complex trade-off between privacy protection and information preservation, where stringent anonymization methods can significantly impact the text's utility for downstream applications. Evaluating the effectiveness of text anonymization proves challenging from both privacy and utility perspectives, as there is no universal benchmark that can comprehensively assess anonymization techniques across diverse, and sometimes contradictory contexts. We present \textsc{tau-eval}, an open-source framework for benchmarking text anonymization methods through the lens of privacy and utility task sensitivity. A Python library\footnote{\faPython~\url{https://pypi.org/project/tau-eval}}, code, documentation and tutorials\footnote{\faGithub~\url{https://github.com/gabrielloiseau/tau-eval} } are publicly available.

\end{abstract}

\section{Introduction}
Privacy protection is a cornerstone of modern legal frameworks, encapsulated in regulations such as the European Union’s General Data Protection Regulation (GDPR) and the United States’ California Consumer Privacy Act (CCPA). These regulations underline the urgency of protecting personal data, particularly in text-based formats—a common medium for sensitive information sharing in domains like healthcare, legal proceedings, and social media. Text anonymization has emerged as a critical tool for this purpose \cite{lison2021anonymisation}, modifying texts to hide identifiable attributes while aiming to preserve their usefulness for downstream applications. However, this process inherently creates a tension: excessive anonymization risks rendering the text unusable for practical tasks, while insufficient redaction leaves private information vulnerable to exposure.

\begin{figure}[ht!]
\centering
  \includegraphics[width=\linewidth,scale=2]{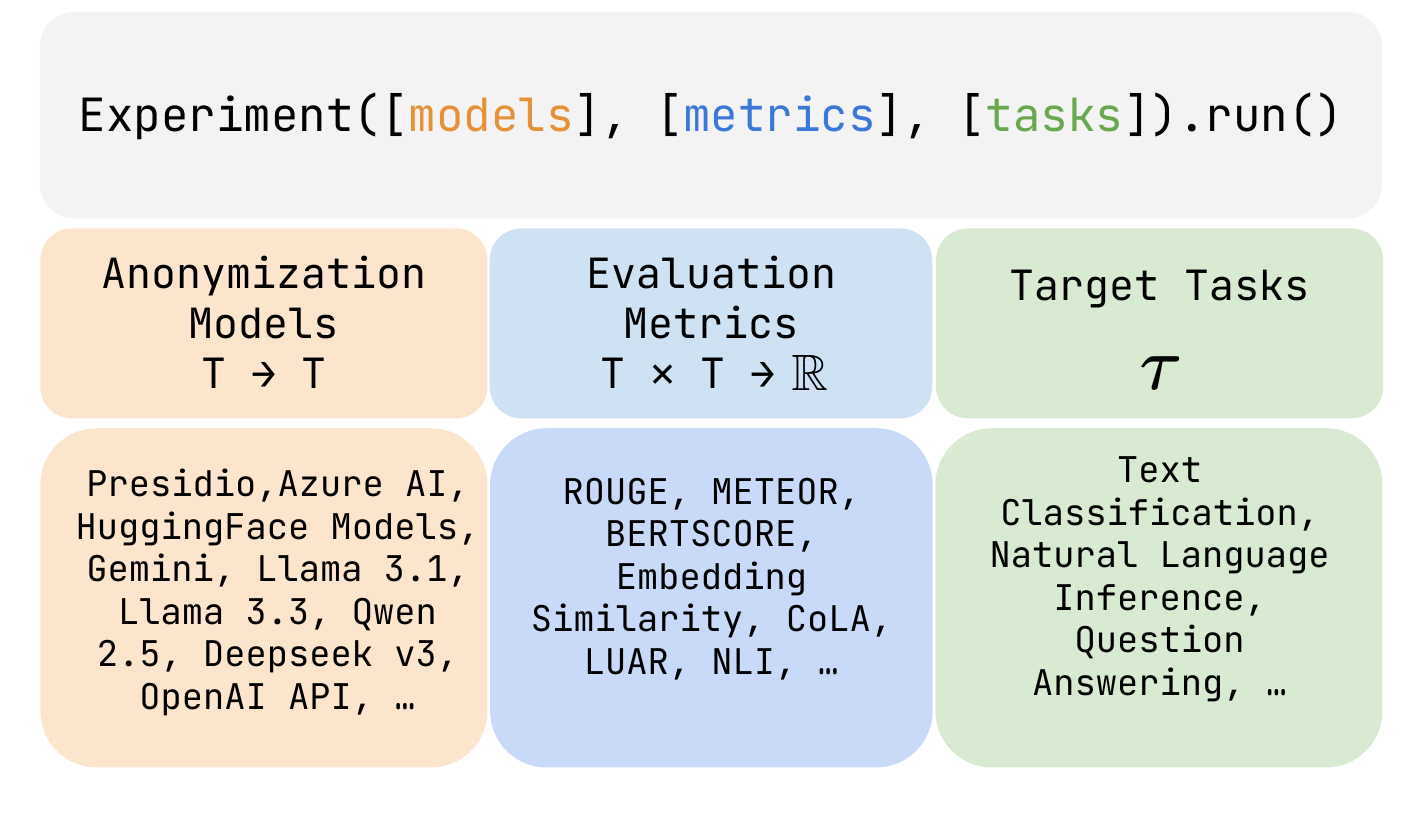}
  \vspace{-0.5cm}
  \caption{Summary of features in \textsc{tau-eval} implementation. Each component can be customized and easily expanded though the system structure. We also include built-in examples for each component. \textsc{tau-eval} relies on a core \textit{Experiment} structure that encapsulate \textit{Models} evaluated on \textit{Tasks} in order to gather \textit{Metrics}.}
  \label{fig:summary}
  \vspace{-0.5cm}
\end{figure}

Current research on text anonymization often prioritizes privacy preservation at the expense of utility, relying on reference-based metrics like ROUGE, BERTScore, or METEOR to measure textual fidelity \cite{staab2024large,pilan-etal-2022-text}. While these metrics assess surface-level content retention and whether the resulting anonymized text lands in the same distribution, they fail to account for the context-dependent utility of anonymized texts \cite{yang2024robustutilitypreservingtextanonymization, loiseau2024tarot}. For instance, a medical report anonymized for public research must retain clinically relevant patterns, whereas a legal document might require syntactic integrity for compliance analysis. Anonymization methods that perform well on generic metrics may strip task-specific features, undermining the value of the data for real-world applications. Consequently, the privacy-utility trade-off remains poorly quantified, leaving practitioners without actionable insights for domain-specific anonymization scenarios \cite{riabi-etal-2024-cloaked}.

To bridge this gap, we introduce \textsc{tau-eval} (\underline{T}ext \underline{A}nonymization \underline{U}tilities \underline{Eval}uation), a framework designed to systematically evaluate both privacy preservation and task-aware utility loss in text anonymization systems. Unlike existing frameworks, \textsc{tau-eval} integrates privacy and utility evaluations across various downstream tasks, enabling granular analysis of how anonymization impacts domain-specific applications. It supports full specification of privacy and utility task targets, which empowers practitioners to evaluate anonymization strategies for their unique requirements and create reproducible evaluation benchmarks.

We showcase \textsc{tau-eval}'s versatility through experiments on two privacy and eight utility tasks. With our framework, we support the development of context-aware anonymization systems that balance privacy with the functional needs of domains like healthcare, social science, and law.

\section{\textsc{Tau-Eval}}
In this section, we present \textsc{tau-eval}'s core design principles, architecture, and the functionality of its main components.

\paragraph{Problem formulation} We formalize text anonymization as a text-to-text transformation task that applies privacy-preserving objectives (e.g., NER-based redaction, authorship obfuscation) to an input text while aiming to preserve its utility. To holistically evaluate anonymization methods, we propose a two-pronged framework: (1) generic metric analysis, measuring surface-level fidelity between original and anonymized texts, and (2) task utility loss quantification, assessing downstream performance degradation caused by anonymization.

Let $\mathcal{D} = \{(x_i,y_i)\}^N_{i=1}$ denote a dataset of $N$ samples, where $x_i$ is an original text and $y_i$ its task-specific label (e.g., classification targets). An anonymization algorithm $\mathcal{A} : x \rightarrow x'$ transforms each $x_i$ into its private counterpart $x'_i$, yielding the anonymized dataset $\mathcal{D}_{\mathrm{priv}}= \{(x'_i,y_i)\}^N_{i=1}$. 
For each sample pair $(x_i,x'_i)$, we compute a similarity metric $s(x_i,x'_i)$ to quantify the text transformation. The overall generic fidelity of $\mathcal{A}$ is then:
\vspace{-2mm}
\begin{align*}
S(\mathcal{A},\mathcal{D}) = \frac{1}{N}\sum^N_{i=1}s(x_i,x'_i),
\end{align*}
providing a task-agnostic measure of anonymization's impact on textual integrity.

To evaluate task-specific utility degradation, we first train a model $f_\theta$ on $\mathcal{D}$, achieving a baseline performance $\mathcal{U}_{\mathrm{orig}}$ on a held-out test set. We then anonymize the test set using $\mathcal{A}$ and evaluate $f_\theta$ on $\mathcal{D}_{\mathrm{priv}}$, obtaining $\mathcal{U}_{\mathrm{priv}}$. This setup mirrors real-world scenarios where users apply anonymization to data before feeding it to a pre-trained task model, which they cannot retrain or modify. The sensitivity for task $\mathcal{T}$ is: $\Delta_\mathcal{T}(\mathcal{A}) = \mathcal{U}_{\mathrm{orig}} - \mathcal{U}_{\mathrm{priv}}$,
where $\Delta_\mathcal{T}$ captures the performance drop attributable to anonymization. Methods with high $S(\mathcal{A},\mathcal{D})$ may still induce significant $\Delta_\mathcal{T}$ if anonymization perturbs task-relevant features, underscoring the necessity of joint analysis. More complex task schemes are also possible, such as training models on partially anonymized datasets to evaluate performance on anonymized outputs \cite{zhai-etal-2022-adversarial}.

\subsection{Design Principles}
\textsc{tau-eval} is guided by three core principles: ease of use, modularity, and customizability, making it a flexible and accessible framework for evaluation in NLP.
\paragraph{\textbf{Ease of use}} \textsc{tau-eval} enables researchers to build complete evaluation pipelines with minimal code. It offers a simple interface and sensible defaults to support rapid development and experimentation.
\paragraph{\textbf{Modularity}} The framework is composed of independent components that can be used selectively. This avoids unnecessary processing and supports a wide range of use cases.
\paragraph{\textbf{Customizability}} Thanks to its modular design, users can easily integrate custom anonymization models, define new features, and implement specialized metrics, allowing for tailored evaluation workflows.

\subsection{Core Elements and Functionalities} 
\textsc{tau-eval} is an open-source Python framework designed to unify and streamline the evaluation of text anonymization systems. By abstracting fragmented experimental workflows into a modular and extensible framework, the library enables researchers to benchmark anonymization models against diverse metrics and tasks with minimal coding effort. Below, we detail its core architecture, integration capabilities, and workflow.
To accommodate the rapid evolution of text generation models and datasets, \textsc{tau-eval} leverages the Hugging Face ecosystem. It natively supports datasets \cite{hf-datasets} within tasks, including local storage and custom preprocessing pipelines, models \cite{hf-transformers}, and evaluations metrics \cite{von-werra-etal-2022-evaluate}. \textsc{tau-eval}’s architecture (Figure~\ref{fig:summary}) revolves around three modular pillars:

\paragraph{Anonymization Models:} The framework treats anonymizers as black-box text-to-text functions. Requiring only a lightweight \texttt{Anonymizer} interface (\texttt{anonymize(text) -> text}). This perspective accommodates a broad spectrum of anonymization techniques; from traditional sequence-to-sequence architectures to modern large-scale language models. Preconfigured templates simplify the use of Hugging Face models (e.g., T5, BART, GPT-2) and LLM APIs via LiteLLM\footnote{\url{https://www.litellm.ai/}}, enabling rapid prototyping.

\paragraph{Evaluation Metrics:} Metrics are designed to capture diverse dimensions of performance. We integrate measures that assess both the retention of semantic content (e.g., overlap-based and semantic similarity scores) and intrinsic properties of anonymized text (e.g., fluency and reference-less evaluations), the framework provides a comprehensive basis for quantifying the efficacy of anonymization methods. Metrics are computed on each pair $(\mathcal{D}, \mathcal{D}_{\mathrm{priv}})$ taken from task datasets. We implement widely used metrics for text anonymization taken from text generation evaluation, such as ROUGE \cite{lin-2004-rouge}, METEOR \cite{banerjee-lavie-2005-meteor}, BERTScore \cite{Zhang2020BERTScore}. We also support sentence-transformers similarity \cite{reimers-gurevych-2019-sentence}, and reference-less metrics based on language models such as CoLA \cite{warstadt-etal-2019-cola}, or Perplexity \cite{jelinek1977perplexity}.

\paragraph{Target Tasks:} Tasks contextualize the evaluation by framing anonymization within downstream applications. We abstract common-use scenarios (e.g., classification, multiple-choice inference) into a unified task paradigm, \textsc{tau-eval} facilitates systematic comparisons across a variety of use cases. We utilize the tasksource ecosystem \cite{sileo-2024-tasksource} to empower rapid access to more than 600 structured datasets and tasks preprocessing with the \texttt{AutoTask} class.
\begin{lstlisting}[language=python]
from tasknet import AutoTask
imdb = AutoTask('imdb')
dynahate = AutoTask('dynahate')
\end{lstlisting}
Using tasknet, we provide a streamlined interface between evaluation tasks and the Hugging Face trainer for efficient fine-tuning. This integration simplifies dataset importing, preprocessing, and model configuration, allowing researchers to concentrate on designing text anonymization pipelines rather than building evaluation frameworks. Additionally, we support additional built-in tasks that don't require model training, such as sensitive entity detection. Table~\ref{tab:implemented} in appendix gives an overview of all features currently integrated into \textsc{tau-eval}.

\section{Usage and Customization}
This section provides implementation details for conducting evaluation experiments and explains how to further customize \textsc{tau-eval} for specific research needs.

\subsection{Running Experiments}
Running an experiment involves three steps: (1) Implement custom anonymizers (see Section~\ref{para:models}) or load preconfigured models present inside \texttt{tau\_eval.models}. (2) Choose from built-in options or add custom metrics and tasks. (3) Instantiate an \texttt{Experiment} object with models, tasks, and metrics. All results are logged for visualization and comparison.

\paragraph{Experiment} The \texttt{Experiment} class takes care of orchestrating the evaluation of each anonymization model. It extracts task information from each task, generates the anonymized task version, and computes each chosen metric, while training relevant models if needed. It relies on \texttt{ExperimentConfig}, a class storing specific user-defined configuration for evaluation.

\paragraph{ExperimentConfig} The \texttt{ExperimentConfig} class provides ways to customize the experiment evaluation. In particular, it allows the user to specify which prediction model to train (from a local model or on the Hugging Face Hub) and fine-tune it on tasks, setting training hyperparameters, or more advanced tasks strategies (should we train the classifier on generated data or not) and logging options.

\texttt{Experiment} results can be stored either as a dictionary variable for immediate use or serialized to a \texttt{.json} file. While the dictionary format is convenient for most direct analysis, JSON serialization helps to store experiment results for versioning and future use. 

\subsection{Customization}
\paragraph{Models} \label{para:models} Anonymization models are defined as a class which allow initialization i.e. model loading and inference with the anonymization method. A new anonymization model can easily be instantiated using this interface:
\begin{lstlisting}[language=Python]
class TestModel(Anonymizer):
  def __init__(self):
    self.name = "Test Model"

  def anonymize(self, text) -> str:
    # Implement anonymization logic

  def anonymize_batch(self, texts: list[str]) -> list[str]:
    # Performs batch anonymization for larger datasets
\end{lstlisting}
\paragraph{Metrics}
Metrics are implemented as functions that accept one or two text inputs and return one or more scores. Experiments can utilize both \textsc{tau-eval}'s built-in metrics and custom functions, provided they follow the signature \texttt{Callable[[str | list[str], str | list[str]], dict]}. The experiment framework automatically detects and handles both built-in and custom metrics.

\paragraph{Tasks}
Tasks enable data loading and classification model integration. While tasksource and tasknet simplify access to the Hugging Face Hub, users can create more sophisticated tasks by implementing the \texttt{CustomTask} interface. This interface requires a \texttt{dataset} attribute containing a Hugging Face dataset and an \texttt{evaluate(self, new\_texts: list[str]) -> dict} method that processes anonymized texts. During experiments, the system anonymizes the task dataset and passes it to the evaluation method.

\subsection{Visualization}
In the evaluation of anonymization systems, \textsc{Tau-Eval} offers a suite of visualization tools designed to streamline the comparative analysis of models across specific tasks. Our system enables to assess performance through one-to-one model comparisons, enhancing interpretability. Additionally, \textsc{Tau-Eval} supports the explicit definition of trade-offs between key metrics (e.g., privacy vs. utility), allowing users to explore the nuanced relationships between competing objectives. To further refine analysis, the framework provides flexible filtering mechanisms, permitting users to isolate experimental results based on models, tasks, or evaluation metrics. We also ensure a more structured and granular assessment of anonymization techniques by serializing parts of the anonymized datasets, which is useful for qualitative analysis and to relaunch experiments without model inference.

\begin{lstlisting}[language=Python, caption=Example running an experiment comparing different de-identification strategies for the MedNLI and \texttt{pii-masking} datasets. BERTScore will be computed on each dataset for each model.]
from tau_eval import Experiment, ExperimentConfig
from tau_eval.models.presidio import (
    UniquePlaceholderPerEntity, 
    EntityDeletion, 
    UniformPlaceholder,
    CategoryPlaceholder,
    FakerPlaceholder,
)
from tau_eval.tasks import DeIdentification
from tasknet import Classification
from datasets import load_dataset

m1 = UniquePlaceholderPerEntity()
m2 = EntityDeletion()
m3 = UniformPlaceholder()
m4 = CategoryPlaceholder()
m5 = FakerPlaceholder()

mednli = Classification(
    dataset=load_dataset("bigbio/mednli"), 
    s1="sentence1", s2="sentence2", y="label"
)
pii = DeIdentification(
    dataset="ai4privacy/pii-masking-400k"
)
config = ExperimentConfig(
    exp_name="test-experiment",
    classifier_name="answerdotai/ModernBERT-base",
    train_task_models=True,
    train_with_generations=False,
)
Experiment(models=[m1,m2,m3,m4,m5],
           metrics=["bertscore"],
           tasks=[mednli, pii],
           config=config
).run()
\end{lstlisting}

\newcommand\rotation{45}
\begin{table*}[!ht]
\centering
\resizebox{0.70\linewidth}{!}{
    \begin{tabular}{lccccccccc}

  \textbf{Model} & \multicolumn{1}{p{2.5ex}}{\rotatebox{\rotation}{\textsc{Privacy}}} & \multicolumn{1}{p{2.5ex}}{\rotatebox{\rotation}{\textsc{IMDB}}} & \multicolumn{1}{p{2.5ex}}{\rotatebox{\rotation}{\textsc{DynaSent}}} & \multicolumn{1}{p{2.5ex}}{\rotatebox{\rotation}{\textsc{Toxicity}}} & \multicolumn{1}{p{2.5ex}}{\rotatebox{\rotation}{\textsc{DynaHate}}} & \multicolumn{1}{p{2.5ex}}{\rotatebox{\rotation}{\textsc{ANLI}}} & \multicolumn{1}{p{2.5ex}}{\rotatebox{\rotation}{\textsc{MedNLI}}} & \multicolumn{1}{p{2.5ex}}{\rotatebox{\rotation}{\textsc{Fraud}}} & \multicolumn{1}{p{2.5ex}}{\rotatebox{\rotation}{\textsc{Fake News}}} \\
\midrule
\emph{PII Redaction}\\
Original & \colorbox{cb-grey}{0} & \CG{95.2}{0}{95.2} & \CG{77.6}{0}{77.6} & \CG{75.6}{0}{75.6} & \CG{83.0}{0}{83.0} & \CG{56.5}{0}{56.5} & \CG{66.3}{0}{66.3} & \CG{99.7}{0}{99.7} & \CG{98.5}{0}{98.5} \\
Presidio & \CG{100}{0}{66.4} & \CG{95.2}{50.0}{95.1} & \CG{77.6}{50.0}{77.5} & \CG{75.6}{33.3}{75.7} & \CG{83.0}{50.0}{80.0} & \CG{56.5}{33.3}{49.6} & \CG{66.3}{33.3}{66.1} & \CG{99.7}{50.8}{97.1} & \CG{98.5}{50.4}{97.9} \\
Gemini-flash-1.5-8b & \CG{100}{0}{96.6} & \CG{95.2}{50.0}{93.7} & \CG{77.6}{33.3}{77.1} & \CG{75.6}{50.0}{56.8} & \CG{83.0}{50.0}{45.0} & \CG{56.5}{33.3}{50.2} & \CG{66.3}{33.3}{65.3} & \CG{99.7}{50.8}{98.3} & \CG{98.5}{50.4}{78.9} \\
Gemini-flash-1.5 & \CG{100}{0}{98.4} & \CG{95.2}{50.0}{94.8} & \CG{77.6}{33.3}{77.4} & \CG{75.6}{50.0}{53.2} & \CG{83.0}{50.0}{41.6} & \CG{56.5}{33.3}{47.8} & \CG{66.3}{33.3}{61.9} & \CG{99.7}{50.8}{97.3} & \CG{98.5}{50.4}{77.1} \\
Llama-3.1-8b & \CG{100}{0}{99.0} & \CG{95.2}{50.0}{94.6} & \CG{77.6}{33.3}{71.8} & \CG{75.6}{50.0}{54.2} & \CG{83.0}{50.0}{49.6} & \CG{56.5}{33.3}{47.7} & \CG{66.3}{33.3}{51.6} & \CG{99.7}{50.8}{92.0} & \CG{98.5}{50.4}{92.7} \\
Llama-3.1-70b & \CG{100}{0}{98.7} & \CG{95.2}{50.0}{95.6} & \CG{77.6}{33.3}{74.6} & \CG{75.6}{50.0}{60.8} & \CG{83.0}{50.0}{51.2} & \CG{56.5}{33.3}{49.2} & \CG{66.3}{33.3}{59.8} & \CG{99.7}{50.8}{98.8} & \CG{98.5}{50.4}{85.2} \\
Llama-3.3-70b & \CG{100}{0}{98.7} & \CG{95.2}{50.0}{95.5} & \CG{77.6}{33.3}{77.0} & \CG{75.6}{50.0}{69.4} & \CG{83.0}{50.0}{53.1} & \CG{56.5}{33.3}{48.9} & \CG{66.3}{33.3}{63.7} & \CG{99.7}{50.8}{98.7} & \CG{98.5}{50.4}{88.9} \\
Qwen-2.5-7b & \CG{100}{0}{95.0} & \CG{95.2}{50.0}{94.9} & \CG{77.6}{33.3}{74.2} & \CG{75.6}{50.0}{65.1} & \CG{83.0}{50.0}{57.5} & \CG{56.5}{33.3}{50.1} & \CG{66.3}{33.3}{61.5} & \CG{99.7}{50.8}{98.6} & \CG{98.5}{50.4}{94.8} \\
Qwen-2.5-72b & \CG{100}{0}{97.8} & \CG{95.2}{50.0}{95.3} & \CG{77.6}{33.3}{76.8} & \CG{75.6}{50.0}{61.1} & \CG{83.0}{50.0}{55.9} & \CG{56.5}{33.3}{52.5} & \CG{66.3}{33.3}{60.4} & \CG{99.7}{50.8}{98.7} & \CG{98.5}{50.4}{88.5} \\
Phi-4 & \CG{100}{0}{97.2} & \CG{95.2}{50.0}{95.1} & \CG{77.6}{33.3}{75.3} & \CG{75.6}{50.0}{58.0} & \CG{83.0}{50.0}{44.7} & \CG{56.5}{33.3}{51.6} & \CG{66.3}{33.3}{61.4} & \CG{99.7}{50.8}{97.5} & \CG{98.5}{50.4}{84.6} \\
\midrule
\emph{Authorship}\\
Original & \colorbox{cb-grey}{0.2} & \CG{95.2}{0}{95.2} & \CG{77.6}{0}{77.6} & \CG{75.6}{0}{75.6} & \CG{83.0}{0}{83.0} & \CG{56.5}{0}{56.5} & \CG{66.3}{0}{66.3} & \CG{99.7}{0}{99.7} & \CG{98.5}{0}{98.5} \\
StyleRemix & \CG{99.8}{10.1}{70.4} & \CG{95.2}{50.0}{82.9} & \CG{77.6}{33.3}{73.2} & \CG{75.6}{50.0}{50.8} & \CG{83.0}{50.0}{47.5} & \CG{56.5}{33.3}{42.5} & \CG{66.3}{33.3}{47.3} & \CG{99.7}{50.8}{97.3} & \CG{98.5}{50.4}{73.9} \\
Gemini-flash-1.5-8b & \CG{99.8}{10.1}{80.6} & \CG{95.2}{50.0}{94.3} & \CG{77.6}{33.3}{70.9} & \CG{75.6}{50.0}{52.4} & \CG{83.0}{50.0}{52.7} & \CG{56.5}{33.3}{45.0} & \CG{66.3}{33.3}{45.9} & \CG{99.7}{50.8}{96.7} & \CG{98.5}{50.4}{47.9} \\
Gemini-flash-1.5 & \CG{99.8}{10.1}{65.6} & \CG{95.2}{50.0}{95.2} & \CG{77.6}{33.3}{71.1} & \CG{75.6}{50.0}{56.9} & \CG{83.0}{50.0}{57.7} & \CG{56.5}{33.3}{44.2} & \CG{66.3}{33.3}{46.2} & \CG{99.7}{50.8}{96.8} & \CG{98.5}{50.4}{58.4} \\
Llama-3.1-8b & \CG{99.8}{10.1}{76.1} & \CG{95.2}{50.0}{93.8} & \CG{77.6}{33.3}{62.6} & \CG{75.6}{50.0}{53.3} & \CG{83.0}{50.0}{55.1} & \CG{56.5}{33.3}{40.7} & \CG{66.3}{33.3}{42.0} & \CG{99.7}{50.8}{94.1} & \CG{98.5}{50.4}{51.5} \\
Llama-3.1-70b & \CG{99.8}{10.1}{69.7} & \CG{95.2}{50.0}{95.2} & \CG{77.6}{33.3}{65.1} & \CG{75.6}{50.0}{51.3} & \CG{83.0}{50.0}{57.7} & \CG{56.5}{33.3}{42.3} & \CG{66.3}{33.3}{41.3} & \CG{99.7}{50.8}{95.3} & \CG{98.5}{50.4}{54.8} \\
Llama-3.3-70b & \CG{99.8}{10.1}{69.3} & \CG{95.2}{50.0}{94.3} & \CG{77.6}{33.3}{65.9} & \CG{75.6}{50.0}{52.9} & \CG{83.0}{50.0}{58.6} & \CG{56.5}{33.3}{41.1} & \CG{66.3}{33.3}{41.8} & \CG{99.7}{50.8}{95.9} & \CG{98.5}{50.4}{60.0} \\
Qwen-2.5-7b & \CG{99.8}{10.1}{66.1} & \CG{95.2}{50.0}{91.9} & \CG{77.6}{33.3}{65.4} & \CG{75.6}{50.0}{58.2} & \CG{83.0}{50.0}{63.5} & \CG{56.5}{33.3}{43.5} & \CG{66.3}{33.3}{46.5} & \CG{99.7}{50.8}{96.2} & \CG{98.5}{50.4}{73.8} \\
Qwen-2.5-72b & \CG{99.8}{10.1}{60.2} & \CG{95.2}{50.0}{94.5} & \CG{77.6}{33.3}{70.7} & \CG{75.6}{50.0}{59.3} & \CG{83.0}{50.0}{66.0} & \CG{56.5}{33.3}{40.3} & \CG{66.3}{33.3}{46.6} & \CG{99.7}{50.8}{96.0} & \CG{98.5}{50.4}{75.7} \\
Phi-4 & \CG{99.8}{10.1}{63.4} & \CG{95.2}{50.0}{94.2} & \CG{77.6}{33.3}{66.4} & \CG{75.6}{50.0}{49.9} & \CG{83.0}{50.0}{59.0} & \CG{56.5}{33.3}{44.4} & \CG{66.3}{33.3}{48.7} & \CG{99.7}{50.8}{95.6} & \CG{98.5}{50.4}{59.6} \\
\midrule
Random & \CG{99.8}{10.1}{10.1} & \CG{95.2}{50.0}{50.0} & \CG{77.6}{33.3}{33.3} & \CG{75.6}{50.0}{50.0} & \CG{83.0}{50.0}{50.0} & \CG{56.5}{33.3}{33.3} & \CG{66.3}{33.3}{33.3} & \CG{99.7}{50.8}{50.8} & \CG{98.5}{50.4}{50.4} \\
\bottomrule
\end{tabular}
}
\caption{Task-specific degradation of each anonymization model (\%). The \textsc{Privacy} column highlights the models performance over the two privacy datasets: \texttt{pii-masking} and IMDB62.}
\label{tab:task_scores}
\end{table*}

\section{Experiments}
To demonstrate the potential of \textsc{Tau-Eval}, we evaluate the utility loss of anonymization methods across two privacy objectives (PII redaction and authorship obfuscation) and eight downstream utility tasks, spanning diverse domains to capture task-sensitive utility loss. Below, we detail our benchmarking setup.

\paragraph{Privacy Tasks}
To quantify privacy risks, we focus on two objectives. First, personally identifiable information (PII) redaction involves automatically identifying and replacing sensitive personal data that could potentially expose individual identities (e.g., names, addresses, etc.). This task is evaluated using the synthetic \texttt{pii-masking} dataset\footnote{\url{https://hf.co/datasets/ai4privacy}}, which generates realistic PII in contextual scenarios while avoiding exposure of real private data. This synthetic approach enables safe benchmarking of anonymizers’ ability to mask sensitive entities without compromising genuine user privacy, we report the masked entity recall as a privacy goal to maximize. Second, authorship obfuscation is a privacy task that tests whether unique authorship features can be used to identify or re-identify an individual's textual content despite anonymization attempts. The task leverages the IMDB62 corpus \cite{seroussi2014authorship}, a widely adopted benchmark for authorship attribution. Here, anonymization aims to obfuscate stylistic fingerprints, testing resilience against authorship inference attacks, we evaluate this task with accuracy scores.

\paragraph{Utility Tasks}
%To measure task-specific utility degradation, we select seven classification tasks spanning diverse domains and linguistic complexity. Sentiment analysis is represented by IMDB movie reviews \cite{imdbdataset} and DynaSent \cite{potts-etal-2020-dynasent}. Toxicity and hate speech detection include Jigsaw Toxic Comments \cite{jigsaw-unintended-bias-in-toxicity-classification} and DynaHate \cite{vidgen-etal-2021-learning}. Specialized domains cover ANLI for adversarial natural language inference \cite{nie2019adversarial}, MedNLI for medical inference \cite{romanov2018lessons}, CLAIR for email fraud detection \cite{radev2008clair}, and FakeNews detection\footnote{\url{https://hf.co/datasets/GonzaloA/fake_news}}. For class-balanced datasets (IMDB, DynaSent, ANLI, MedNLI, FRAUD, FakeNews), we report accuracy; for imbalanced tasks (Toxicity, DynaHate), F1 scores are used. All tasks are fine-tuned on ModernBERT-large \cite{warner2024modernbert} (hyperparameters reported in Appendix~\ref{apx:hyperparameters}), selected for its balance of efficiency and state-of-the-art NLP performance.
To comprehensively measure task-specific utility degradation, we select eight classification tasks spanning diverse domains and linguistic complexity, each chosen to probe distinct challenges in privacy-utility trade-offs. IMDB movie reviews \cite{imdbdataset} and the adversarially constructed DynaSent dataset \cite{potts-etal-2020-dynasent} serve as baselines for \textit{sentiment analysis}. \textit{Toxicity and hate speech detection} (Jigsaw Toxic Comments \cite{jigsaw-unintended-bias-in-toxicity-classification}, DynaHate \cite{vidgen-etal-2021-learning}) evaluate anonymization’s impact on socially critical tasks, where over-redaction may obscure harmful language. ANLI \cite{nie2019adversarial}, an adversarial \textit{natural language inference} (NLI) dataset, assesses whether anonymization preserves logical consistency between premises and hypotheses. MedNLI \cite{romanov2018lessons} addresses the medical domain’s acute privacy needs, where anonymization must protect patient identities without distorting diagnostic inferences. 
CLAIR \cite{radev2008clair} (\textit{fraudulent email detection}) and FakeNews detection\footnote{\url{https://hf.co/datasets/GonzaloA/fake_news}} examines anonymization’s effect on stylistic and structural cues critical for identifying malicious intent and misinformation. We add further details for each task in Appendix~\ref{apx:datasets}.

For class-balanced datasets (IMDB, DynaSent, ANLI, MedNLI, FRAUD, FakeNews), we report accuracy; for imbalanced tasks (Toxicity, DynaHate), F1 scores are used. All tasks are fine-tuned on ModernBERT-large \cite{warner2024modernbert}, selected for its balance of efficiency and state-of-the-art NLP performance. We also include a random classifier predicting the distribution of target labels as a baseline.

\begin{figure*}[ht]
    \centering
    \includegraphics[width=0.85\linewidth]{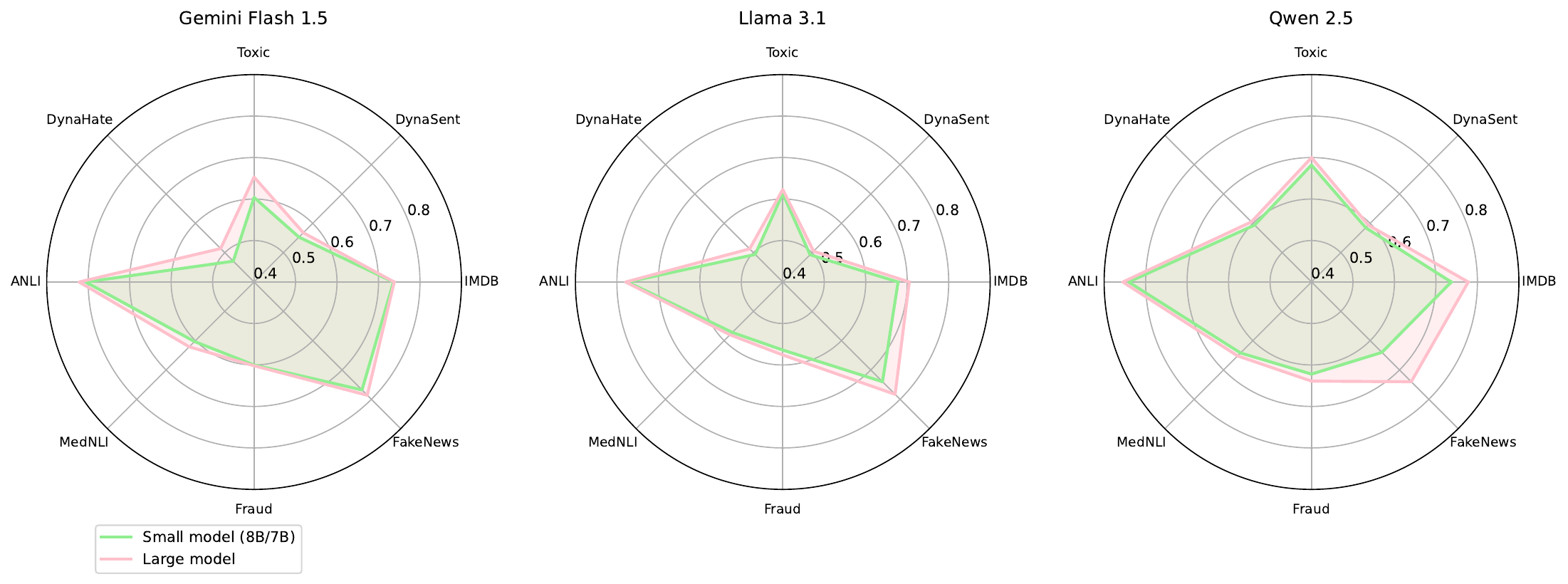}
    \caption{SBERT cosine similarity difference between each tested model and its smaller version.}
    \label{fig:sbert-result}
\end{figure*}

\paragraph{Anonymization Models}
We benchmark two families of anonymization methods, each of them corresponding to one or the other privacy task. For PII redaction, we evaluate the NER-based Presidio anonymizer \cite{MsPresidio} as a baseline and compare it against four state-of-the-art LLMs: Gemini Flash 1.5 \cite{geminiteam2024gemini15unlockingmultimodal}, Llama-3.1/3.3 \cite{grattafiori2024llama3herdmodels}, Qwen-2.5 \cite{qwen2025qwen25technicalreport}, and Phi-4 \cite{abdin2024phi4technicalreport} prompted for entity replacement. For authorship obfuscation, we test StyleRemix \cite{fisher-etal-2024-styleremix}, a state-of-the-art baseline, alongside the same LLMs repurposed with style-transfer prompts to disrupt stylistic cues while retaining task-relevant content. Model size effects are analyzed by comparing parameter variants (e.g., Llama-3.1-8B vs. 70B). We access instruction-tuned language models through the OpenRouter API.

\section{Results}

Table~\ref{tab:task_scores} summarizes the performance of anonymization methods across privacy and utility tasks. LLMs emerge as the most effective anonymizers for both PII redaction and authorship obfuscation, achieving best privacy protection. However, this comes at a pronounced cost to utility. Toxicity and hate detection tasks exhibit the steepest performance drops (up to $42\%$ utility decrease for Gemini-flash-1.5 on DynaHate), likely due to LLMs’ fine-tuning safeguards against reproducing harmful content. Notably, in the authorship setting, model size inversely correlates with privacy and utility: smaller models tend to do more content modifications to the original text (see Figure~\ref{fig:sbert-result}), leading to better privacy and worse utility. Utility degradation varies dramatically across tasks. Sentiment analysis tasks like IMDB and DynaSent exhibit resilience to anonymization, whereas ANLI, MedNLI and fake news detection degrade significantly. Crucially, no single anonymization method dominates across all tasks, but each familly of models exibit similar performances accross tasks.

We also performed a correlation analysis between general purpose metrics commonly used in state-of-the-art for utility preservation and task-specific performance measures, computing Kendall's Tau correlation scores as reported in Table~\ref{tab:correlation}. The results strongly emphasize the limitations of generic metrics in reliably predicting task utility across different domains. No single metric consistently demonstrates strong correlation with actual utility loss across the evaluated tasks. The sole exception to this pattern was observed in the MedNLI task, where moderate to strong correlations were found. This anomaly can be potentially attributed to the relatively short text size in MedNLI samples, which may result in a more direct relationship between surface-level text modifications and task performance.

\begin{table}[t]
    \centering
    %\vspace{-3mm}
    \resizebox{\columnwidth}{!}{
    \begin{tabular}{lccccc}
        \toprule
        Task & BERTScore & METEOR & ROUGE & SBERT & CoLA \\
        \midrule
        IMDB & 0.349 & 0.304 & 0.349 & 0.096 & 0.141 \\
        DynaSent & 0.735 & 0.691 & 0.705 & 0.735 & 0.573 \\
        Tox & 0.681 & 0.726 & 0.681 & 0.516 & 0.576 \\
        DynaHate & 0.007 & 0.051 & 0.095 & 0.317 & 0.051 \\
        ANLI & 0.627 & 0.573 & 0.720 & 0.352 & 0.544 \\
        MedNLI & 0.867 & 0.764 & 0.838 & 0.882 & 0.705 \\
        Fraud & 0.518 & 0.464 & 0.464 & 0.597 & 0.420 \\
        FakeNews & 0.699 & 0.759 & 0.774 & 0.248 & 0.323 \\
        \midrule
        Average & 0.560 & 0.542 & 0.578 & 0.468 & 0.417 \\
        \bottomrule
    \end{tabular}}
    \caption{Correlation of different metrics with task performance for all utility tasks.}
    \vspace{-5mm}
    \label{tab:correlation}
\end{table}

\section{Related Work}
Most existing research addressing similar challenges has primarily focused on evaluating model robustness against sentence-level perturbations. Tools like TextFlint \cite{wang-etal-2021-textflint} offer diverse text noising methods, but emphasize evaluation of downstream task models rather than analyzing the noise mechanisms themselves. PrivKit \cite{10.1145/3626232.3653284} represents a notable exception in its attempt to establish a comprehensive privacy-utility evaluation pipeline for heterogeneous data types. However, it was primarily designed for location and facial data rather than natural language processing applications and consequently lacks the unified framework necessary to support the diverse requirements of NLP evaluation tasks. Furthermore, PrivKit operates as a standalone system without integration capabilities for popular machine learning frameworks, limiting its accessibility and adoption in existing workflows.

\section{Conclusion}
In this paper we introduce \textsc{Tau-Eval}, an evaluation framework enabling unified, reproducible benchmarking of text anonymization systems across both privacy risks and task-specific utility degradation. We study the task sensitivity of language models instructed for anonymization on eight downstream tasks and justify its relevance as a complement to task-agnostic metrics. Our experiments confirm the complexity of achieving a clear trade-off, revealing that no single method dominates across tasks.

\section*{Limitations}
While our framework advances the evaluation of task-sensitive anonymization, several limitations highlight directions for future work. First, 
%our privacy evaluation assumes passive adversaries attempting to infer sensitive attributes (e.g., PII, authorship) from anonymized text. This scope excludes active adversarial scenarios where attackers exploit knowledge of the anonymization algorithm; for example, training on partially anonymized data to reverse-engineer privacy mechanisms, as explored in adversarial threat models \cite{zhai-etal-2022-adversarial}. Incorporating such attacks, which mirror real-world data reconstruction risks, would strengthen privacy robustness assessments. % The same technique could also be applied for utility to measure the re-training impact of classifier with anonymized texts. We include this option in the framework implementation and plan to study its consequence 
our study operates within a monolingual (English) context. Multilingual anonymization introduces unique challenges \cite{riabi-etal-2024-cloaked}. Low-resource languages further compound these issues due to sparse PII detection tools and limited benchmark datasets, restricting the generalizability of findings to global contexts.

Second, utility loss measurements depend on the choice of task classifier. While we present evaluations using ModernBERT-large for cross-task comparability, domain-specific architectures (e.g., ClinicalBERT \cite{Huang2019ClinicalBERTMC}, CamemBERT-bio \cite{touchent-de-la-clergerie-2024-camembert-bio}) or multilingual models (e.g., XLM-T \cite{barbieri-etal-2022-xlm}) may yield divergent utility trade-offs. For instance, medical anonymization evaluated via a clinically fine-tuned model could reveal subtler impacts on diagnostic inference than our general-purpose setup captures. 

Finally, we did implement LLMs mainly using privacy-focused prompts, and could have implemented utility-specific LLMs that perform anonymization when given specific downstream task requirements. Such task-aware anonymization would fundamentally shift the privacy-utility trade-off by limiting data usage to predefined tasks. This presents an interesting direction for future research.

\section*{Ethics and Broader Impact Statement}
While \textsc{Tau-Eval} facilitates the benchmarking of anonymization methods, it does not itself perform anonymization or guarantee privacy. Users are responsible for interpreting results in context and for deploying anonymization strategies that align with applicable legal, ethical, and domain-specific standards. We emphasize that \textsc{Tau-Eval} should not be used as a substitute for legal compliance (e.g., GDPR) but as a research tool to assist in privacy-aware NLP development.

\bibliography{ref}

\begin{thebibliography}{53}
\providecommand{\natexlab}[1]{#1}

\bibitem[{Abdin et~al.(2024)Abdin, Aneja, Behl, Bubeck, Eldan, Gunasekar, Harrison, Hewett, Javaheripi, Kauffmann, Lee, Lee, Li, Liu, Mendes, Nguyen, Price, de~Rosa, Saarikivi, Salim, Shah, Wang, Ward, Wu, Yu, Zhang, and Zhang}]{abdin2024phi4technicalreport}
Marah Abdin, Jyoti Aneja, Harkirat Behl, Sébastien Bubeck, Ronen Eldan, Suriya Gunasekar, Michael Harrison, Russell~J. Hewett, Mojan Javaheripi, Piero Kauffmann, James~R. Lee, Yin~Tat Lee, Yuanzhi Li, Weishung Liu, Caio C.~T. Mendes, Anh Nguyen, Eric Price, Gustavo de~Rosa, Olli Saarikivi, Adil Salim, Shital Shah, Xin Wang, Rachel Ward, Yue Wu, Dingli Yu, Cyril Zhang, and Yi~Zhang. 2024.
\newblock \href {https://arxiv.org/abs/2412.08905} {Phi-4 technical report}.
\newblock \emph{Preprint}, arXiv:2412.08905.

\bibitem[{Banerjee and Lavie(2005)}]{banerjee-lavie-2005-meteor}
Satanjeev Banerjee and Alon Lavie. 2005.
\newblock \href {https://aclanthology.org/W05-0909/} {{METEOR}: An automatic metric for {MT} evaluation with improved correlation with human judgments}.
\newblock In \emph{Proceedings of the {ACL} Workshop on Intrinsic and Extrinsic Evaluation Measures for Machine Translation and/or Summarization}, pages 65--72, Ann Arbor, Michigan. Association for Computational Linguistics.

\bibitem[{Bao and Carpuat(2024)}]{bao-carpuat-2024-kip}
Calvin Bao and Marine Carpuat. 2024.
\newblock \href {https://doi.org/10.18653/v1/2024.naacl-long.480} {{K}eep it {P}rivate: Unsupervised privatization of online text}.
\newblock In \emph{Proceedings of the 2024 Conference of the North American Chapter of the Association for Computational Linguistics: Human Language Technologies (Volume 1: Long Papers)}, pages 8678--8693, Mexico City, Mexico. Association for Computational Linguistics.

\bibitem[{Barbieri et~al.(2022)Barbieri, Espinosa~Anke, and Camacho-Collados}]{barbieri-etal-2022-xlm}
Francesco Barbieri, Luis Espinosa~Anke, and Jose Camacho-Collados. 2022.
\newblock \href {https://aclanthology.org/2022.lrec-1.27/} {{XLM}-{T}: Multilingual language models in {T}witter for sentiment analysis and beyond}.
\newblock In \emph{Proceedings of the Thirteenth Language Resources and Evaluation Conference}, pages 258--266, Marseille, France. European Language Resources Association.

\bibitem[{Ben Cheikh~Larbi et~al.(2023)Ben Cheikh~Larbi, Burchardt, and Roller}]{ben-cheikh-larbi-etal-2023-clinical}
Iyadh Ben Cheikh~Larbi, Aljoscha Burchardt, and Roland Roller. 2023.
\newblock \href {https://doi.org/10.18653/v1/2023.eacl-srw.11} {Clinical text anonymization, its influence on downstream {NLP} tasks and the risk of re-identification}.
\newblock In \emph{Proceedings of the 17th Conference of the European Chapter of the Association for Computational Linguistics: Student Research Workshop}, pages 105--111, Dubrovnik, Croatia. Association for Computational Linguistics.

\bibitem[{Berg et~al.(2020)Berg, Henriksson, and Dalianis}]{berg-etal-2020-impact}
Hanna Berg, Aron Henriksson, and Hercules Dalianis. 2020.
\newblock \href {https://doi.org/10.18653/v1/2020.louhi-1.1} {The impact of de-identification on downstream named entity recognition in clinical text}.
\newblock In \emph{Proceedings of the 11th International Workshop on Health Text Mining and Information Analysis}, pages 1--11, Online. Association for Computational Linguistics.

\bibitem[{cjadams et~al.(2019)cjadams, Borkan, inversion, Sorensen, Dixon, Vasserman, and nithum}]{jigsaw-unintended-bias-in-toxicity-classification}
cjadams, Daniel Borkan, inversion, Jeffrey Sorensen, Lucas Dixon, Lucy Vasserman, and nithum. 2019.
\newblock \href {https://kaggle.com/competitions/jigsaw-unintended-bias-in-toxicity-classification} {Jigsaw unintended bias in toxicity classification}.
\newblock Kaggle.

\bibitem[{Cunha et~al.(2024)Cunha, Duarte, Andrade, Mendes, and Vilela}]{10.1145/3626232.3653284}
Mariana Cunha, Guilherme Duarte, Ricardo Andrade, Ricardo Mendes, and Jo\~{a}o~P. Vilela. 2024.
\newblock \href {https://doi.org/10.1145/3626232.3653284} {Privkit: A toolkit of privacy-preserving mechanisms for heterogeneous data types}.
\newblock In \emph{Proceedings of the Fourteenth ACM Conference on Data and Application Security and Privacy}, CODASPY '24, page 319–324, New York, NY, USA. Association for Computing Machinery.

\bibitem[{Fan et~al.(2021)Fan, Bhosale, Schwenk, Ma, El-Kishky, Goyal, Baines, Celebi, Wenzek, Chaudhary, Goyal, Birch, Liptchinsky, Edunov, Grave, Auli, and Joulin}]{10.5555/3546258.3546365}
Angela Fan, Shruti Bhosale, Holger Schwenk, Zhiyi Ma, Ahmed El-Kishky, Siddharth Goyal, Mandeep Baines, Onur Celebi, Guillaume Wenzek, Vishrav Chaudhary, Naman Goyal, Tom Birch, Vitaliy Liptchinsky, Sergey Edunov, Edouard Grave, Michael Auli, and Armand Joulin. 2021.
\newblock Beyond english-centric multilingual machine translation.
\newblock \emph{J. Mach. Learn. Res.}, 22(1).

\bibitem[{Fisher et~al.(2024)Fisher, Hallinan, Lu, Gordon, Harchaoui, and Choi}]{fisher-etal-2024-styleremix}
Jillian Fisher, Skyler Hallinan, Ximing Lu, Mitchell~L Gordon, Zaid Harchaoui, and Yejin Choi. 2024.
\newblock \href {https://doi.org/10.18653/v1/2024.emnlp-main.241} {{S}tyle{R}emix: Interpretable authorship obfuscation via distillation and perturbation of style elements}.
\newblock In \emph{Proceedings of the 2024 Conference on Empirical Methods in Natural Language Processing}, pages 4172--4206, Miami, Florida, USA. Association for Computational Linguistics.

\bibitem[{Grattafiori et~al.(2024)Grattafiori, Dubey, Jauhri, Pandey, Kadian, and et~al.}]{grattafiori2024llama3herdmodels}
Aaron Grattafiori, Abhimanyu Dubey, Abhinav Jauhri, Abhinav Pandey, Abhishek Kadian, and Ahmad Al-Dahle et~al. 2024.
\newblock \href {https://arxiv.org/abs/2407.21783} {The llama 3 herd of models}.
\newblock \emph{Preprint}, arXiv:2407.21783.

\bibitem[{Huang et~al.(2019)Huang, Altosaar, and Ranganath}]{Huang2019ClinicalBERTMC}
Kexin Huang, Jaan Altosaar, and Rajesh Ranganath. 2019.
\newblock \href {https://api.semanticscholar.org/CorpusID:119308351} {Clinicalbert: Modeling clinical notes and predicting hospital readmission}.
\newblock \emph{ArXiv}, abs/1904.05342.

\bibitem[{Jelinek et~al.(1977)Jelinek, Mercer, Bahl, and Baker}]{jelinek1977perplexity}
Fred Jelinek, Robert~L Mercer, Lalit~R Bahl, and James~K Baker. 1977.
\newblock Perplexity—a measure of the difficulty of speech recognition tasks.
\newblock \emph{The Journal of the Acoustical Society of America}, 62(S1):S63--S63.

\bibitem[{Kandula et~al.(2024)Kandula, Karakos, Qiu, and Ulicny}]{kandula-etal-2024-improving}
Hemanth Kandula, Damianos Karakos, Haoling Qiu, and Brian Ulicny. 2024.
\newblock \href {https://aclanthology.org/2024.privatenlp-1.14/} {Improving authorship privacy: Adaptive obfuscation with the dynamic selection of techniques}.
\newblock In \emph{Proceedings of the Fifth Workshop on Privacy in Natural Language Processing}, pages 137--142, Bangkok, Thailand. Association for Computational Linguistics.

\bibitem[{Kiela et~al.(2021)Kiela, Bartolo, Nie, Kaushik, Geiger, Wu, Vidgen, Prasad, Singh, Ringshia, Ma, Thrush, Riedel, Waseem, Stenetorp, Jia, Bansal, Potts, and Williams}]{kiela-etal-2021-dynabench}
Douwe Kiela, Max Bartolo, Yixin Nie, Divyansh Kaushik, Atticus Geiger, Zhengxuan Wu, Bertie Vidgen, Grusha Prasad, Amanpreet Singh, Pratik Ringshia, Zhiyi Ma, Tristan Thrush, Sebastian Riedel, Zeerak Waseem, Pontus Stenetorp, Robin Jia, Mohit Bansal, Christopher Potts, and Adina Williams. 2021.
\newblock \href {https://doi.org/10.18653/v1/2021.naacl-main.324} {Dynabench: Rethinking benchmarking in {NLP}}.
\newblock In \emph{Proceedings of the 2021 Conference of the North American Chapter of the Association for Computational Linguistics: Human Language Technologies}, pages 4110--4124, Online. Association for Computational Linguistics.

\bibitem[{Laban et~al.(2021)Laban, Schnabel, Bennett, and Hearst}]{laban-etal-2021-keep}
Philippe Laban, Tobias Schnabel, Paul Bennett, and Marti~A. Hearst. 2021.
\newblock \href {https://doi.org/10.18653/v1/2021.acl-long.498} {Keep it simple: Unsupervised simplification of multi-paragraph text}.
\newblock In \emph{Proceedings of the 59th Annual Meeting of the Association for Computational Linguistics and the 11th International Joint Conference on Natural Language Processing (Volume 1: Long Papers)}, pages 6365--6378, Online. Association for Computational Linguistics.

\bibitem[{Lampoltshammer et~al.(2019)Lampoltshammer, Thurnay, and Eibl}]{Lampoltshammer}
Thomas Lampoltshammer, Lőrinc Thurnay, and Gregor Eibl. 2019.
\newblock \href {https://doi.org/10.1007/978-3-658-27495-5_5} {\emph{Impact of Anonymization on Sentiment Analysis of Twitter Postings}}, pages 41--48.

\bibitem[{Lange et~al.(2020)Lange, Adel, and Str{\"o}tgen}]{lange-etal-2020-closing}
Lukas Lange, Heike Adel, and Jannik Str{\"o}tgen. 2020.
\newblock \href {https://doi.org/10.18653/v1/2020.acl-main.621} {Closing the gap: Joint de-identification and concept extraction in the clinical domain}.
\newblock In \emph{Proceedings of the 58th Annual Meeting of the Association for Computational Linguistics}, pages 6945--6952, Online. Association for Computational Linguistics.

\bibitem[{Lhoest et~al.(2021)Lhoest, Villanova~del Moral, Jernite, Thakur, von Platen, Patil, Chaumond, Drame, Plu, Tunstall, Davison, {\v{S}}a{\v{s}}ko, Chhablani, Malik, Brandeis, Le~Scao, Sanh, Xu, Patry, McMillan-Major, Schmid, Gugger, Delangue, Matussi{\`e}re, Debut, Bekman, Cistac, Goehringer, Mustar, Lagunas, Rush, and Wolf}]{hf-datasets}
Quentin Lhoest, Albert Villanova~del Moral, Yacine Jernite, Abhishek Thakur, Patrick von Platen, Suraj Patil, Julien Chaumond, Mariama Drame, Julien Plu, Lewis Tunstall, Joe Davison, Mario {\v{S}}a{\v{s}}ko, Gunjan Chhablani, Bhavitvya Malik, Simon Brandeis, Teven Le~Scao, Victor Sanh, Canwen Xu, Nicolas Patry, Angelina McMillan-Major, Philipp Schmid, Sylvain Gugger, Cl{\'e}ment Delangue, Th{\'e}o Matussi{\`e}re, Lysandre Debut, Stas Bekman, Pierric Cistac, Thibault Goehringer, Victor Mustar, Fran{\c{c}}ois Lagunas, Alexander Rush, and Thomas Wolf. 2021.
\newblock \href {https://doi.org/10.18653/v1/2021.emnlp-demo.21} {Datasets: A community library for natural language processing}.
\newblock In \emph{Proceedings of the 2021 Conference on Empirical Methods in Natural Language Processing: System Demonstrations}, pages 175--184, Online and Punta Cana, Dominican Republic. Association for Computational Linguistics.

\bibitem[{Lin(2004)}]{lin-2004-rouge}
Chin-Yew Lin. 2004.
\newblock \href {https://aclanthology.org/W04-1013/} {{ROUGE}: A package for automatic evaluation of summaries}.
\newblock In \emph{Text Summarization Branches Out}, pages 74--81, Barcelona, Spain. Association for Computational Linguistics.

\bibitem[{Lison et~al.(2021)Lison, Pil{\'a}n, S{\'a}nchez, Batet, and {\O}vrelid}]{lison2021anonymisation}
Pierre Lison, Ildik{\'o} Pil{\'a}n, David S{\'a}nchez, Montserrat Batet, and Lilja {\O}vrelid. 2021.
\newblock Anonymisation models for text data: State of the art, challenges and future directions.
\newblock In \emph{Proceedings of the 59th Annual Meeting of the Association for Computational Linguistics and the 11th International Joint Conference on Natural Language Processing (Volume 1: Long Papers)}, pages 4188--4203.

\bibitem[{Liu et~al.(2022)Liu, Swayamdipta, Smith, and Choi}]{liu-etal-2022-wanli}
Alisa Liu, Swabha Swayamdipta, Noah~A. Smith, and Yejin Choi. 2022.
\newblock \href {https://doi.org/10.18653/v1/2022.findings-emnlp.508} {{WANLI}: Worker and {AI} collaboration for natural language inference dataset creation}.
\newblock In \emph{Findings of the Association for Computational Linguistics: EMNLP 2022}, pages 6826--6847, Abu Dhabi, United Arab Emirates. Association for Computational Linguistics.

\bibitem[{Loiseau et~al.(2025)Loiseau, Sileo, Riquet, Meyer, and Tommasi}]{loiseau2024tarot}
Gabriel Loiseau, Damien Sileo, Damien Riquet, Maxime Meyer, and Marc Tommasi. 2025.
\newblock \href {https://doi.org/10.18653/v1/2025.privatenlp-main.2} {{TAROT}: Task-oriented authorship obfuscation using policy optimization methods}.
\newblock In \emph{Proceedings of the Sixth Workshop on Privacy in Natural Language Processing}, pages 14--31, Albuquerque, New Mexico. Association for Computational Linguistics.

\bibitem[{Lothritz et~al.(2023)Lothritz, Lebichot, Allix, Ezzini, Bissyand{\'e}, Klein, Boytsov, Lefebvre, and Goujon}]{lothritz-etal-2023-evaluating}
Cedric Lothritz, Bertrand Lebichot, Kevin Allix, Saad Ezzini, Tegawend{\'e} Bissyand{\'e}, Jacques Klein, Andrey Boytsov, Cl{\'e}ment Lefebvre, and Anne Goujon. 2023.
\newblock \href {https://aclanthology.org/2023.nodalida-1.2/} {Evaluating the impact of text de-identification on downstream {NLP} tasks}.
\newblock In \emph{Proceedings of the 24th Nordic Conference on Computational Linguistics (NoDaLiDa)}, pages 10--16, T{\'o}rshavn, Faroe Islands. University of Tartu Library.

\bibitem[{Maas et~al.(2011)Maas, Daly, Pham, Huang, Ng, and Potts}]{imdbdataset}
Andrew~L. Maas, Raymond~E. Daly, Peter~T. Pham, Dan Huang, Andrew~Y. Ng, and Christopher Potts. 2011.
\newblock \href {http://www.aclweb.org/anthology/P11-1015} {Learning word vectors for sentiment analysis}.
\newblock In \emph{Proceedings of the 49th Annual Meeting of the Association for Computational Linguistics: Human Language Technologies}, pages 142--150, Portland, Oregon, USA. Association for Computational Linguistics.

\bibitem[{Mendels et~al.(2018)Mendels, Peled, Vaisman~Levy, Hart, Rosenthal, Lahiani et~al.}]{MsPresidio}
Omri Mendels, Coby Peled, Nava Vaisman~Levy, Sharon Hart, Tomer Rosenthal, Limor Lahiani, et~al. 2018.
\newblock \href {https://microsoft.github.io/presidio} {{Microsoft Presidio}: Context aware, pluggable and customizable pii anonymization service for text and images}.

\bibitem[{Nie et~al.(2020)Nie, Williams, Dinan, Bansal, Weston, and Kiela}]{nie2019adversarial}
Yixin Nie, Adina Williams, Emily Dinan, Mohit Bansal, Jason Weston, and Douwe Kiela. 2020.
\newblock Adversarial nli: A new benchmark for natural language understanding.
\newblock In \emph{Proceedings of the 58th Annual Meeting of the Association for Computational Linguistics}. Association for Computational Linguistics.

\bibitem[{Patsakis and Lykousas(2023)}]{patsakis2023manvsmachinestruggle}
Constantinos Patsakis and Nikolaos Lykousas. 2023.
\newblock Man vs the machine in the struggle for effective text anonymisation in the age of large language models.
\newblock \emph{Scientific Reports}, 13(1):16026.

\bibitem[{Pil{\'a}n et~al.(2022)Pil{\'a}n, Lison, {\O}vrelid, Papadopoulou, S{\'a}nchez, and Batet}]{pilan-etal-2022-text}
Ildik{\'o} Pil{\'a}n, Pierre Lison, Lilja {\O}vrelid, Anthi Papadopoulou, David S{\'a}nchez, and Montserrat Batet. 2022.
\newblock \href {https://doi.org/10.1162/coli_a_00458} {The text anonymization benchmark ({TAB}): A dedicated corpus and evaluation framework for text anonymization}.
\newblock \emph{Computational Linguistics}, 48(4):1053--1101.

\bibitem[{Potts et~al.(2021)Potts, Wu, Geiger, and Kiela}]{potts-etal-2020-dynasent}
Christopher Potts, Zhengxuan Wu, Atticus Geiger, and Douwe Kiela. 2021.
\newblock \href {https://doi.org/10.18653/v1/2021.acl-long.186} {{D}yna{S}ent: A dynamic benchmark for sentiment analysis}.
\newblock In \emph{Proceedings of the 59th Annual Meeting of the Association for Computational Linguistics and the 11th International Joint Conference on Natural Language Processing (Volume 1: Long Papers)}, pages 2388--2404, Online. Association for Computational Linguistics.

\bibitem[{Qwen et~al.(2025)Qwen, :, Yang, Yang, Zhang, Hui, Zheng, Yu, Li, Liu, Huang, Wei, Lin, Yang, Tu, Zhang, Yang, Yang, Zhou, Lin, Dang, Lu, Bao, Yang, Yu, Li, Xue, Zhang, Zhu, Men, Lin, Li, Tang, Xia, Ren, Ren, Fan, Su, Zhang, Wan, Liu, Cui, Zhang, and Qiu}]{qwen2025qwen25technicalreport}
Qwen, :, An~Yang, Baosong Yang, Beichen Zhang, Binyuan Hui, Bo~Zheng, Bowen Yu, Chengyuan Li, Dayiheng Liu, Fei Huang, Haoran Wei, Huan Lin, Jian Yang, Jianhong Tu, Jianwei Zhang, Jianxin Yang, Jiaxi Yang, Jingren Zhou, Junyang Lin, Kai Dang, Keming Lu, Keqin Bao, Kexin Yang, Le~Yu, Mei Li, Mingfeng Xue, Pei Zhang, Qin Zhu, Rui Men, Runji Lin, Tianhao Li, Tianyi Tang, Tingyu Xia, Xingzhang Ren, Xuancheng Ren, Yang Fan, Yang Su, Yichang Zhang, Yu~Wan, Yuqiong Liu, Zeyu Cui, Zhenru Zhang, and Zihan Qiu. 2025.
\newblock \href {https://arxiv.org/abs/2412.15115} {Qwen2.5 technical report}.
\newblock \emph{Preprint}, arXiv:2412.15115.

\bibitem[{Radev(2008)}]{radev2008clair}
Dragomir Radev. 2008.
\newblock \href {http://aclweb.org/aclwiki} {Clair collection of fraud email}.
\newblock ACL Data and Code Repository, ADCR2008T001.

\bibitem[{Reimers and Gurevych(2019)}]{reimers-gurevych-2019-sentence}
Nils Reimers and Iryna Gurevych. 2019.
\newblock \href {https://doi.org/10.18653/v1/D19-1410} {Sentence-{BERT}: Sentence embeddings using {S}iamese {BERT}-networks}.
\newblock In \emph{Proceedings of the 2019 Conference on Empirical Methods in Natural Language Processing and the 9th International Joint Conference on Natural Language Processing (EMNLP-IJCNLP)}, pages 3982--3992, Hong Kong, China. Association for Computational Linguistics.

\bibitem[{Riabi et~al.(2024)Riabi, Mahamdi, Mouilleron, and Seddah}]{riabi-etal-2024-cloaked}
Arij Riabi, Menel Mahamdi, Virginie Mouilleron, and Djam{\'e} Seddah. 2024.
\newblock \href {https://aclanthology.org/2024.privatenlp-1.13/} {Cloaked classifiers: Pseudonymization strategies on sensitive classification tasks}.
\newblock In \emph{Proceedings of the Fifth Workshop on Privacy in Natural Language Processing}, pages 123--136, Bangkok, Thailand. Association for Computational Linguistics.

\bibitem[{Rivera-Soto et~al.(2021)Rivera-Soto, Miano, Ordonez, Chen, Khan, Bishop, and Andrews}]{rivera-soto-etal-2021-learning}
Rafael~A. Rivera-Soto, Olivia~Elizabeth Miano, Juanita Ordonez, Barry~Y. Chen, Aleem Khan, Marcus Bishop, and Nicholas Andrews. 2021.
\newblock \href {https://doi.org/10.18653/v1/2021.emnlp-main.70} {Learning universal authorship representations}.
\newblock In \emph{Proceedings of the 2021 Conference on Empirical Methods in Natural Language Processing}, pages 913--919, Online and Punta Cana, Dominican Republic. Association for Computational Linguistics.

\bibitem[{Romanov and Shivade(2018)}]{romanov2018lessons}
Alexey Romanov and Chaitanya Shivade. 2018.
\newblock \href {https://doi.org/10.18653/v1/D18-1187} {Lessons from natural language inference in the clinical domain}.
\newblock In \emph{Proceedings of the 2018 Conference on Empirical Methods in Natural Language Processing}, pages 1586--1596, Brussels, Belgium. Association for Computational Linguistics.

\bibitem[{Seroussi et~al.(2014)Seroussi, Zukerman, and Bohnert}]{seroussi2014authorship}
Yanir Seroussi, Ingrid Zukerman, and Fabian Bohnert. 2014.
\newblock Authorship attribution with topic models.
\newblock \emph{Computational Linguistics}, 40(2):269--310.

\bibitem[{Sileo(2024)}]{sileo-2024-tasksource}
Damien Sileo. 2024.
\newblock \href {https://aclanthology.org/2024.lrec-main.1361} {tasksource: A large collection of {NLP} tasks with a structured dataset preprocessing framework}.
\newblock In \emph{Proceedings of the 2024 Joint International Conference on Computational Linguistics, Language Resources and Evaluation (LREC-COLING 2024)}, pages 15655--15684, Torino, Italia. ELRA and ICCL.

\bibitem[{Staab et~al.(2024{\natexlab{a}})Staab, Vero, Balunovic, and Vechev}]{staab2024large}
Robin Staab, Mark Vero, Mislav Balunovic, and Martin Vechev. 2024{\natexlab{a}}.
\newblock Large language models are anonymizers.
\newblock In \emph{ICLR 2024 Workshop on Reliable and Responsible Foundation Models}.

\bibitem[{Staab et~al.(2024{\natexlab{b}})Staab, Vero, Balunović, and Vechev}]{staab24beyond}
Robin Staab, Mark Vero, Mislav Balunović, and Martin Vechev. 2024{\natexlab{b}}.
\newblock Beyond memorization: Violating privacy via inference with large language models.
\newblock In \emph{The Twelfth International Conference on Learning Representations}.

\bibitem[{Team et~al.(2024)Team, Georgiev, Lei, Burnell, and et~al.}]{geminiteam2024gemini15unlockingmultimodal}
Gemini Team, Petko Georgiev, Ving~Ian Lei, Ryan Burnell, and Libin~Bai et~al. 2024.
\newblock \href {https://arxiv.org/abs/2403.05530} {Gemini 1.5: Unlocking multimodal understanding across millions of tokens of context}.
\newblock \emph{Preprint}, arXiv:2403.05530.

\bibitem[{Touchent and de~la Clergerie(2024)}]{touchent-de-la-clergerie-2024-camembert-bio}
Rian Touchent and {\'E}ric de~la Clergerie. 2024.
\newblock \href {https://aclanthology.org/2024.lrec-main.241} {{C}amem{BERT}-bio: Leveraging continual pre-training for cost-effective models on {F}rench biomedical data}.
\newblock In \emph{Proceedings of the 2024 Joint International Conference on Computational Linguistics, Language Resources and Evaluation (LREC-COLING 2024)}, pages 2692--2701, Torino, Italia. ELRA and ICCL.

\bibitem[{Vidgen et~al.(2021)Vidgen, Thrush, Waseem, and Kiela}]{vidgen-etal-2021-learning}
Bertie Vidgen, Tristan Thrush, Zeerak Waseem, and Douwe Kiela. 2021.
\newblock \href {https://doi.org/10.18653/v1/2021.acl-long.132} {Learning from the worst: Dynamically generated datasets to improve online hate detection}.
\newblock In \emph{Proceedings of the 59th Annual Meeting of the Association for Computational Linguistics and the 11th International Joint Conference on Natural Language Processing (Volume 1: Long Papers)}, pages 1667--1682, Online. Association for Computational Linguistics.

\bibitem[{Von~Werra et~al.(2022)Von~Werra, Tunstall, Thakur, Luccioni, Thrush, Piktus, Marty, Rajani, Mustar, and Ngo}]{von-werra-etal-2022-evaluate}
Leandro Von~Werra, Lewis Tunstall, Abhishek Thakur, Sasha Luccioni, Tristan Thrush, Aleksandra Piktus, Felix Marty, Nazneen Rajani, Victor Mustar, and Helen Ngo. 2022.
\newblock \href {https://doi.org/10.18653/v1/2022.emnlp-demos.13} {Evaluate {\&} evaluation on the hub: Better best practices for data and model measurements}.
\newblock In \emph{Proceedings of the 2022 Conference on Empirical Methods in Natural Language Processing: System Demonstrations}, pages 128--136, Abu Dhabi, UAE. Association for Computational Linguistics.

\bibitem[{Wang et~al.(2021)Wang, Liu, Gui, Zhang, Zou, Zhou, Ye, Zhang, Zheng, Pang, Wu, Li, Zhang, Ma, Fei, Cai, Zhao, Hu, Yan, Tan, Hu, Bian, Liu, Qin, Zhu, Xing, Fu, Zhang, Peng, Zheng, Zhou, Wei, Qiu, and Huang}]{wang-etal-2021-textflint}
Xiao Wang, Qin Liu, Tao Gui, Qi~Zhang, Yicheng Zou, Xin Zhou, Jiacheng Ye, Yongxin Zhang, Rui Zheng, Zexiong Pang, Qinzhuo Wu, Zhengyan Li, Chong Zhang, Ruotian Ma, Zichu Fei, Ruijian Cai, Jun Zhao, Xingwu Hu, Zhiheng Yan, Yiding Tan, Yuan Hu, Qiyuan Bian, Zhihua Liu, Shan Qin, Bolin Zhu, Xiaoyu Xing, Jinlan Fu, Yue Zhang, Minlong Peng, Xiaoqing Zheng, Yaqian Zhou, Zhongyu Wei, Xipeng Qiu, and Xuanjing Huang. 2021.
\newblock \href {https://doi.org/10.18653/v1/2021.acl-demo.41} {{T}ext{F}lint: Unified multilingual robustness evaluation toolkit for natural language processing}.
\newblock In \emph{Proceedings of the 59th Annual Meeting of the Association for Computational Linguistics and the 11th International Joint Conference on Natural Language Processing: System Demonstrations}, pages 347--355, Online. Association for Computational Linguistics.

\bibitem[{Warner et~al.(2025)Warner, Chaffin, Clavi{\'e}, Weller, Hallstr{\"o}m, Taghadouini, Gallagher, Biswas, Ladhak, Aarsen, Adams, Howard, and Poli}]{warner2024modernbert}
Benjamin Warner, Antoine Chaffin, Benjamin Clavi{\'e}, Orion Weller, Oskar Hallstr{\"o}m, Said Taghadouini, Alexis Gallagher, Raja Biswas, Faisal Ladhak, Tom Aarsen, Griffin~Thomas Adams, Jeremy Howard, and Iacopo Poli. 2025.
\newblock \href {https://doi.org/10.18653/v1/2025.acl-long.127} {Smarter, better, faster, longer: A modern bidirectional encoder for fast, memory efficient, and long context finetuning and inference}.
\newblock In \emph{Proceedings of the 63rd Annual Meeting of the Association for Computational Linguistics (Volume 1: Long Papers)}, pages 2526--2547, Vienna, Austria. Association for Computational Linguistics.

\bibitem[{Warstadt et~al.(2019)Warstadt, Singh, and Bowman}]{warstadt-etal-2019-cola}
Alex Warstadt, Amanpreet Singh, and Samuel~R. Bowman. 2019.
\newblock \href {https://doi.org/10.1162/tacl_a_00290} {Neural network acceptability judgments}.
\newblock \emph{Transactions of the Association for Computational Linguistics}, 7:625--641.

\bibitem[{Wolf et~al.(2020)Wolf, Debut, Sanh, Chaumond, Delangue, Moi, Cistac, Rault, Louf, Funtowicz, Davison, Shleifer, von Platen, Ma, Jernite, Plu, Xu, Le~Scao, Gugger, Drame, Lhoest, and Rush}]{hf-transformers}
Thomas Wolf, Lysandre Debut, Victor Sanh, Julien Chaumond, Clement Delangue, Anthony Moi, Pierric Cistac, Tim Rault, Remi Louf, Morgan Funtowicz, Joe Davison, Sam Shleifer, Patrick von Platen, Clara Ma, Yacine Jernite, Julien Plu, Canwen Xu, Teven Le~Scao, Sylvain Gugger, Mariama Drame, Quentin Lhoest, and Alexander Rush. 2020.
\newblock \href {https://doi.org/10.18653/v1/2020.emnlp-demos.6} {Transformers: State-of-the-art natural language processing}.
\newblock In \emph{Proceedings of the 2020 Conference on Empirical Methods in Natural Language Processing: System Demonstrations}, pages 38--45, Online. Association for Computational Linguistics.

\bibitem[{Xing et~al.(2024)Xing, Venkatraman, Le, and Lee}]{xing2024alison}
Eric Xing, Saranya Venkatraman, Thai Le, and Dongwon Lee. 2024.
\newblock Alison: Fast and effective stylometric authorship obfuscation.
\newblock In \emph{AAAI}.

\bibitem[{Yang et~al.(2025)Yang, Zhu, and Gurevych}]{yang2024robustutilitypreservingtextanonymization}
Tianyu Yang, Xiaodan Zhu, and Iryna Gurevych. 2025.
\newblock \href {https://doi.org/10.18653/v1/2025.acl-long.1404} {Robust utility-preserving text anonymization based on large language models}.
\newblock In \emph{Proceedings of the 63rd Annual Meeting of the Association for Computational Linguistics (Volume 1: Long Papers)}, pages 28922--28941, Vienna, Austria. Association for Computational Linguistics.

\bibitem[{Zhai et~al.(2022)Zhai, Rusert, Shafiq, and Srinivasan}]{zhai-etal-2022-adversarial}
Wanyue Zhai, Jonathan Rusert, Zubair Shafiq, and Padmini Srinivasan. 2022.
\newblock \href {https://doi.org/10.18653/v1/2022.acl-long.509} {Adversarial authorship attribution for deobfuscation}.
\newblock In \emph{Proceedings of the 60th Annual Meeting of the Association for Computational Linguistics (Volume 1: Long Papers)}, pages 7372--7384, Dublin, Ireland. Association for Computational Linguistics.

\bibitem[{Zhang et~al.(2020{\natexlab{a}})Zhang, Zhao, Saleh, and Liu}]{zhang2019pegasus}
Jingqing Zhang, Yao Zhao, Mohammad Saleh, and Peter~J. Liu. 2020{\natexlab{a}}.
\newblock Pegasus: pre-training with extracted gap-sentences for abstractive summarization.
\newblock In \emph{Proceedings of the 37th International Conference on Machine Learning}, ICML'20. JMLR.org.

\bibitem[{Zhang et~al.(2020{\natexlab{b}})Zhang, Kishore, Wu, Weinberger, and Artzi}]{Zhang2020BERTScore}
Tianyi Zhang, Varsha Kishore, Felix Wu, Kilian~Q. Weinberger, and Yoav Artzi. 2020{\natexlab{b}}.
\newblock \href {https://openreview.net/forum?id=SkeHuCVFDr} {Bertscore: Evaluating text generation with bert}.
\newblock In \emph{International Conference on Learning Representations}.

\end{thebibliography}

\appendix

\section{Datasets \& Task Details}
\label{apx:datasets}
Our evaluation framework employs datasets that span privacy preservation and task-specific utility, carefully selected to reflect real-world anonymization challenges across diverse domains. Below, we elaborate on their structure, relevance, and role in quantifying the privacy-utility trade-off.
\paragraph{pii-masking} This synthetic dataset simulates real-world privacy risks by generating text records containing 63 classes of personally identifiable information (PII) (see \url{https://hf.co/ai4privacy}), including names, addresses, medical IDs, and financial data. Unlike real-world corpora, synthetic generation avoids ethical concerns while enabling controlled benchmarking of anonymizers’ ability to mask sensitive entities. We sample 5,000 English texts from its 400k corpus, ensuring representation of all PII categories. The dataset’s structured noise injection (e.g., realistic address formats, contextualized medical terms) tests anonymizers’ precision in distinguishing sensitive from non-sensitive content which is a critical capability for GDPR/CCPA compliance. 
\paragraph{IMDB62} Derived from the IMDb movie review corpus \cite{seroussi2014authorship}, this dataset contains 62,000 texts from 62 distinct authors, equally distributed. We focus on a 10-class subset (first 10 authors) to evaluate authorship obfuscation. The task challenges anonymizers to disrupt stylistic fingerprints (e.g., syntactic patterns, lexical preferences). 
\paragraph{IMDB} \cite{imdbdataset}: A classic binary sentiment classification task (positive/negative) on 50k movie reviews. Its balanced distribution and informal language (e.g., user-generated reviews) test anonymization’s impact on opinion-driven text utility.
\paragraph{DynaSent}  \cite{potts-etal-2020-dynasent}: A ternary sentiment dataset (positive/neutral/negative) constructed via the Dynabench platform \cite{kiela-etal-2021-dynabench}, where adversarial examples are iteratively generated to exploit model weaknesses. DynaSent’s complexity evaluates whether anonymization exacerbates or mitigates robustness to subtle linguistic perturbations.
\paragraph{Toxicity} \cite{jigsaw-unintended-bias-in-toxicity-classification}: A binary classification task identifying toxic language (“rude, disrespectful, or disruptive” content) in online comments. The dataset’s inherent class imbalance ($\approx 10\%$ toxic) tests anonymization’s impact on minority class performance and harmful content detection, a critical consideration for moderation systems.
\paragraph{DynaHate} \cite{vidgen-etal-2021-learning}: Built using Dynabench’s adversarial framework, this dataset targets hate speech detection with examples designed to bypass automated filters. Its adversarial nature stresses anonymization’s ability to preserve subtle hate indicators (e.g., dog whistles, coded language) while removing identifiers.
\paragraph{ANLI} \cite{nie2019adversarial}: An adversarial natural language inference (NLI) dataset where premises and hypotheses are iteratively refined to challenge model reasoning. By anonymizing both text spans, we test consistency in logical inference: a measure of whether anonymization disrupts semantic relationships critical for NLI. 
\paragraph{MedNLI} \cite{romanov2018lessons}: A medical NLI dataset derived from clinical notes, requiring models to infer entailment/contradiction relationships between patient descriptions and hypotheses. Anonymization here risks altering clinically relevant entities (e.g., medications, symptoms), making it a benchmark for domain-specific utility preservation.
\paragraph{Fraud} \cite{radev2008clair}: A corpus of fraudulent emails spanning 1988–present, where anonymization must balance redacting PII (e.g., names, account numbers) with preserving stylistic cues (e.g., urgency markers, grammatical errors) indicative of fraud.
\paragraph{Fake News}: A binary classification task on 45k news articles (sourced from \url{https://hf.co/datasets/GonzaloA/fake_news}) that evaluates anonymization’s impact on semantic coherence in misinformation contexts. Articles often contain named entities (e.g., politicians, organizations) whose anonymization may alter the perceived credibility of claims.

\section{Motivations from Related Work}
This section presents a concise overview of the background and related work on evaluating the utility of text anonymization, providing the foundation for the motivations of this paper.

Prior work on text anonymization utility has been most extensive in the medical domain \cite{lange-etal-2020-closing, ben-cheikh-larbi-etal-2023-clinical}, where protecting Protected Health Information (PHI) is ethically and legally required. Studies show that excessive anonymization can reduce clinical value \cite{berg-etal-2020-impact}. Outside the medical domain, utility assessments are narrower, often limited to sentiment analysis \cite{yang2024robustutilitypreservingtextanonymization, patsakis2023manvsmachinestruggle, Lampoltshammer}.
 \citet{lothritz-etal-2023-evaluating} provide a broader evaluation focused on GLUE tasks \cite{lothritz-etal-2023-evaluating} but target name redaction only, overlooking semantic and syntactic effects.
Recent findings highlight that pseudonymization impacts tasks differently \cite{riabi-etal-2024-cloaked}, motivating systematic, multi-task evaluation.

Other domains, such as authorship anonymization, emphasize identity protection and utility is often assessed narrowly through text generation metrics like n-gram preservation and fluency \cite{xing2024alison, fisher-etal-2024-styleremix, kandula-etal-2024-improving, bao-carpuat-2024-kip}. The rise of large language models as both privacy attackers \cite{staab24beyond} and anonymizers \cite{staab2024large} underscores the need for comprehensive evaluation protocols that combine task-agnostic and task-specific measures.

\begin{table}[!ht]
\centering
\renewcommand{\arraystretch}{1.2}
\begin{tabular}{p{0.9\columnwidth}}
\hline
\textbf{Already Implemented Models} \\
\hline
Pseudonymization \cite{riabi-etal-2024-cloaked} \\
-- Entity Deletion \\
-- Uniform Placeholder \\
-- Category-Specific Placeholder \\
-- Unique Placeholder per Entity \\
-- Unique Substitute per Entity \\
Authorship Obfuscation \\
-- StyleRemix \cite{fisher-etal-2024-styleremix}\\
-- Simplification \cite{laban-etal-2021-keep}\\
-- Back Translation \cite{10.5555/3546258.3546365}\\
-- Paraphrasing \cite{zhang2019pegasus}\\
$[+]$ \textit{preconfigured templates for HF models \& API-based LLMs} \\
\hline
\textbf{Already Implemented Metrics} \\
\hline
Reference-based \\
-- ROUGE \cite{lin-2004-rouge}\\
-- METEOR \cite{banerjee-lavie-2005-meteor}\\
-- BERTScore \cite{Zhang2020BERTScore}\\
-- NLI \cite{liu-etal-2022-wanli}\\
-- LUAR similarity \cite{rivera-soto-etal-2021-learning}\\
-- SBERT similarity \cite{reimers-gurevych-2019-sentence}\\
Reference-less \\
-- CoLA \cite{warstadt-etal-2019-cola}\\
-- Perplexity \cite{jelinek1977perplexity}\\
$[+]$ \textit{preconfigured templates for HF evaluate metrics} \\
\hline
\textbf{Already Implemented Tasks} \\
\hline
-- De-identification \\
-- IMDB Authorship Attribution \\
-- Text Anonymization Benchmark \cite{pilan-etal-2022-text} \\
-- Sequence Classification \\
-- Token Classification \\
-- Multiple Choice \\
-- Sequence-to-Sequence \\
$[+]$ \textit{preconfigured templates for tasksource tasks} \\
\hline
\end{tabular}
\caption{Overview of already implemented models, metrics, and tasks at the time of submission (July 2025).}
\label{tab:implemented}
\end{table}

\section{Prompts}
\label{apx:prompts}
Figures~\ref{fig:author-prompt} and \ref{fig:pii-prompt} show the prompts used for authorship privacy and PII redaction tasks. 
\begin{figure}[h!]
\improvedParbox{
    \textbf{User:}
    \texttt{Rewrite the text to significantly alter its style, tone, and word choice while preserving the original meaning. Use figurative or descriptive language, vary sentence structures, adjust tone (e.g., formal to conversational), and employ unique synonyms. Avoid retaining distinctive stylistic markers. Respond only with the transformed text.}
}
\caption{Authorship obfuscation prompt}
\label{fig:author-prompt}
\end{figure}

\begin{figure}[h!]
\improvedParbox{
    \textbf{User:}
    \texttt{Anonymize user-submitted text by replacing personally identifiable information (PII) like names, dates, locations, and professions with alternate values. Ensure the rewritten text is natural, coherent, and contextually consistent while preserving the original meaning and tone. Respond only with the transformed text.}
}
\caption{PII redaction prompt}
\label{fig:pii-prompt}
\end{figure}
\end{document}